\documentclass{article}
\usepackage{ijcai19}

\usepackage{times}
\usepackage{soul}
\usepackage{url}
\usepackage[hidelinks]{hyperref}
\usepackage[utf8]{inputenc}
\usepackage[small]{caption}
\usepackage{graphicx}
\usepackage{amsmath}
\usepackage{booktabs}
\usepackage{subfig}
\usepackage{algorithm}
\usepackage{algpseudocode}
\usepackage{multirow}

\title{Autoencoding Binary Classifiers for Supervised Anomaly Detection}

\author{
Yuki Yamanaka$^1$\footnote{Contact Author}\and
Tomoharu Iwata$^2$\and
Hiroshi Takahashi$^3$\and \\
Masanori Yamada$^1$\And
Sekitoshi Kanai$^3$\\
\affiliations
$^1$ NTT Secure Platform Laboratories\\
$^2$ NTT Communication Science Laboratories\\
$^3$ NTT Software Innovation Center\\
\emails
\{yuuki.yamanaka.kb,
tomoharu.iwata.gy,
hiroshi.takahashi.bm,
masanori.yamada.cm,
sekitoshi.kanai.fu
\}@hco.ntt.co.jp
}

\begin{document}

\maketitle

\begin{abstract}
  We propose the Autoencoding Binary Classifiers (ABC), a novel supervised anomaly detector based on the Autoencoder (AE).
  There are two main approaches in anomaly detection: supervised and unsupervised.
  The supervised approach accurately detects the known anomalies included in training data, but it cannot detect the unknown anomalies.
  % The supervised approach can accurately discriminate known anomalies, which are included in the training dataset, from normal data points.
  % Nevertheless, it cannot detect unknown anomalies not included in the dataset.
  Meanwhile, the unsupervised approach can detect both known and unknown anomalies that are located away from normal data points.
  However, it does not detect known anomalies as accurately as the supervised approach.
  Furthermore, even if we have labeled normal data points and anomalies, the unsupervised approach cannot utilize these labels.
  The ABC is a probabilistic binary classifier that effectively exploits the label information, where normal data points are modeled using the AE as a component.
  By maximizing the likelihood, the AE in the proposed ABC is trained to minimize the reconstruction error for normal data points, and to maximize it for known anomalies.
  Since our approach becomes able to reconstruct the normal data points accurately and fails to reconstruct the known and unknown anomalies, it can accurately discriminate both known and unknown anomalies from normal data points.
  Experimental results show that the ABC achieves higher detection performance than existing supervised and unsupervised methods.
\end{abstract}

\section{Introduction} %%%%%%%%%%%%%%%%%%%%%%%%%%%%%%%%%%%%%%%%%%%%%%%%%%%%%%%

\begin{figure}[t]
  \begin{center}
    \subfloat[Supervised]{
      \includegraphics[width=25truemm]{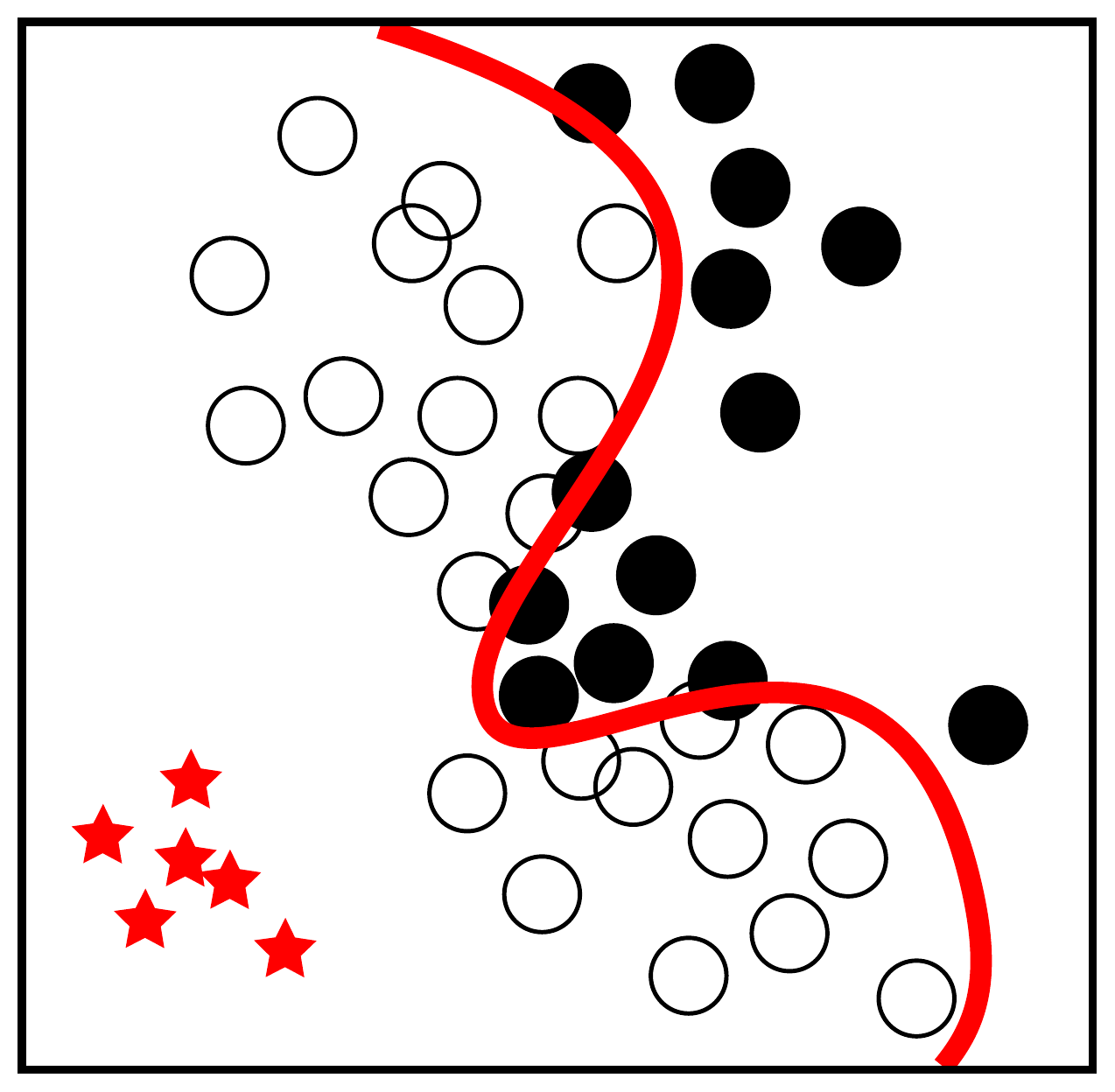}
    }
    \subfloat[Unsupervised]{
      \includegraphics[width=25truemm]{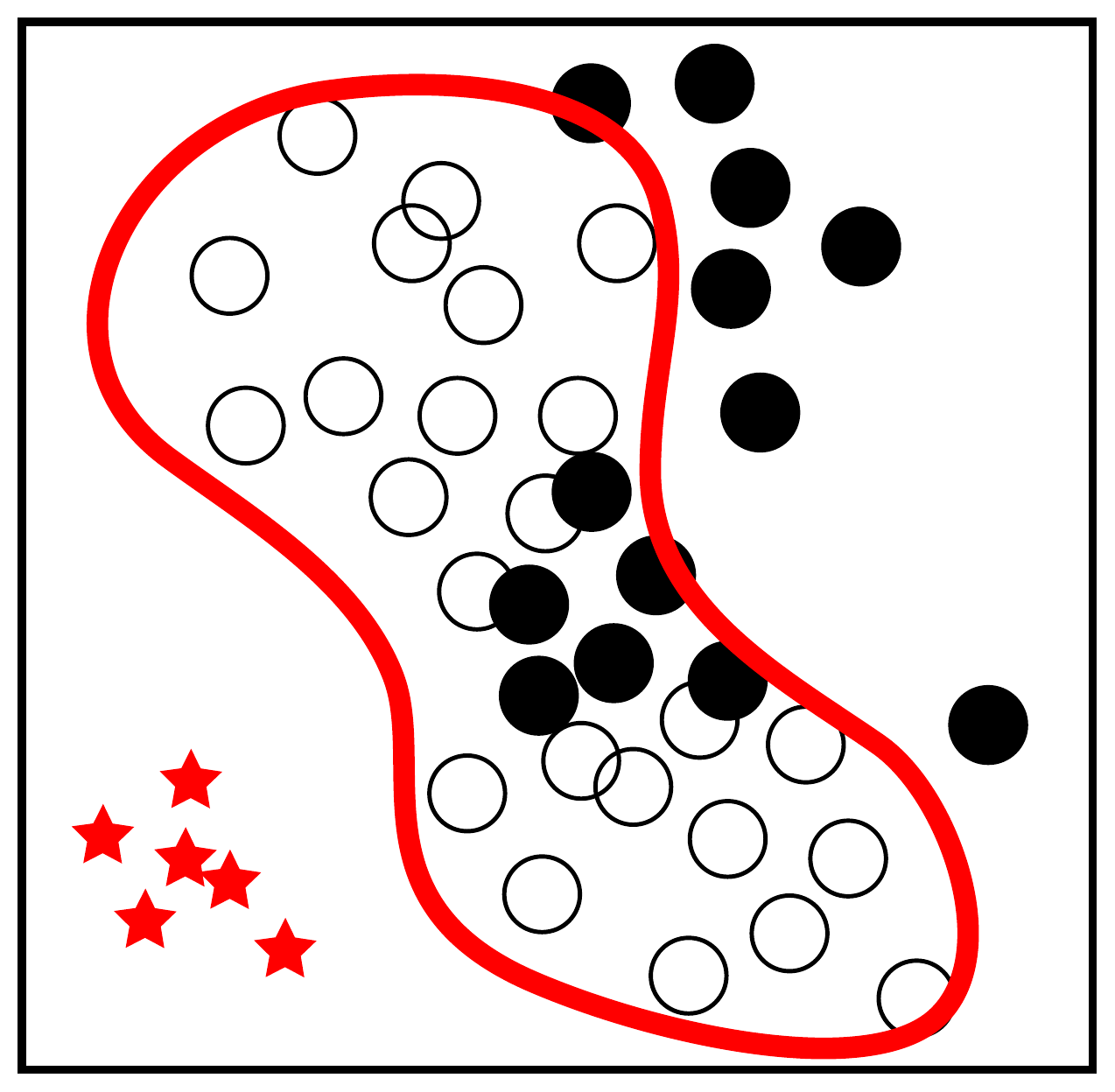}
    }
    \subfloat[Our approach]{
      \includegraphics[width=25truemm]{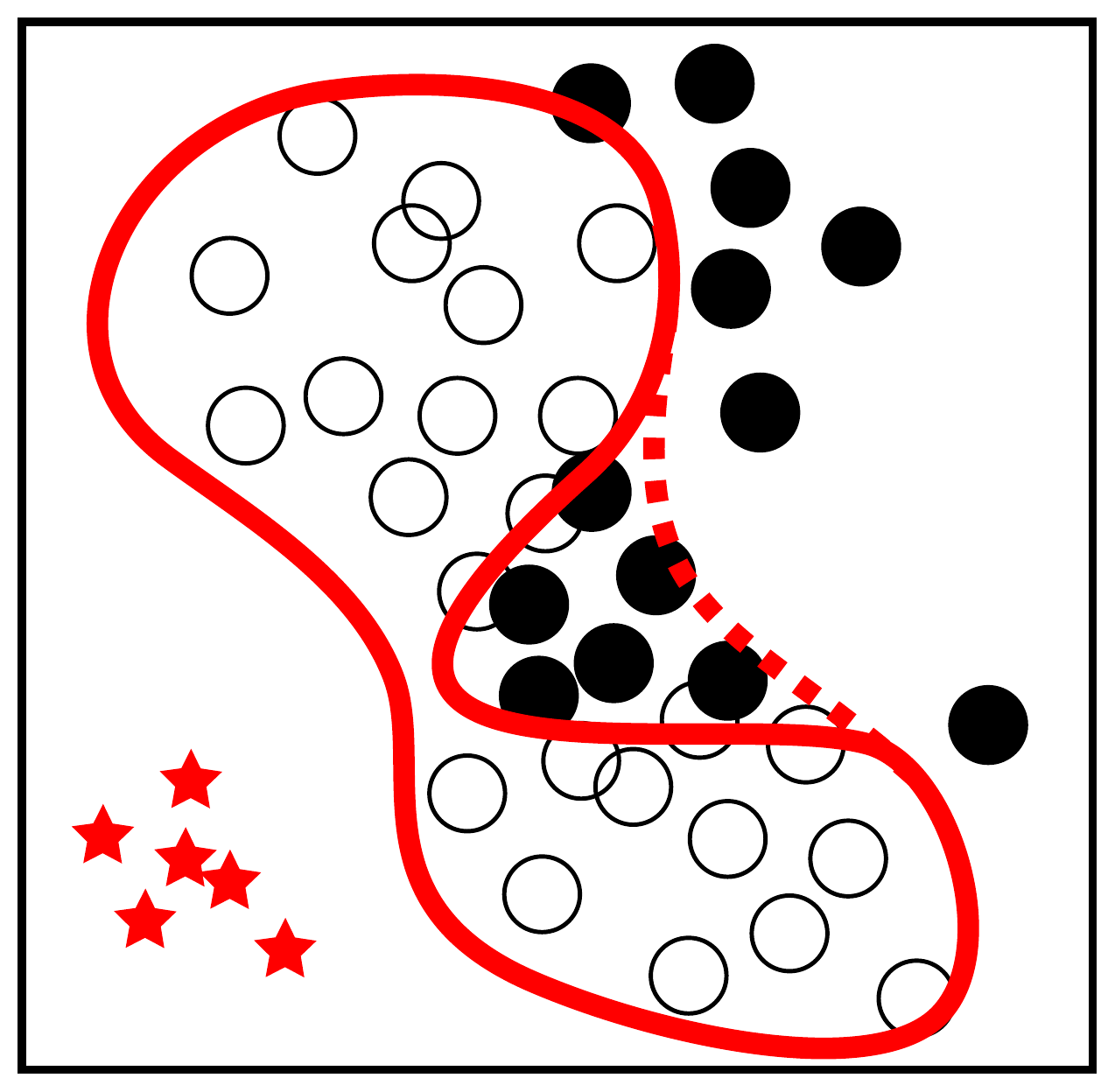}
    }
  \caption{
  (a) Supervised, (b) unsupervised and (c) our approaches.
  White circles, black circles, and red stars represent normal data points, known anomalies, and unknown anomalies, respectively.
  Red lines show the decision boundary.
  % (a) Image of supervised approaches.
  % Although these approaches can detect known anomalies, they fail to detect unknown anomalies.
  % (b) Image of unsupervised approaches.
  % Although these approaches can detect anomalies located away from normal data points, they fail to detect anomalies located near the normal data points.
  % (c) Image of the LRC and our approach.
  % This approach can detect both known and unknown anomalies accurately.
  }
  \label{fig:1}
  \end{center}
\end{figure}

% 異常検知は大事だよ
Anomaly detection is one of the most important tasks in artificial intelligence and is widely performed in many areas,
such as cyber security \cite{kwon2017survey}, complex system management \cite{liu2008isolation}, and material inspection \cite{tagawa2015structured}.

% 教師あり手法はこんなとこがいいけど、こんなとこがダメ。
Anomaly detection methods can be categorized into supervised and unsupervised approaches.
A supervised approach learns a classification rule that discriminates anomalies from normal data points \cite{nasrabadi2007pattern}.
Figure \ref{fig:1}(a) shows the image of the supervised approach.
This approach utilizes the label information, which indicates whether each data point is an anomaly or not,
and can achieve high detection performance for known anomalies included in a dataset.
However, a supervised approach is effective only for limited applications for two reasons.
First, it cannot detect the unknown anomalies not included in the dataset.
Generally, it is difficult to obtain a training dataset that covers all types of possible anomalies in advance.
For example, in malware detection, unknown anomalies will appear one after another.
% In such cases, a supervised approach cannot detect these unknown anomalies.
Second, the standard supervised approaches do not work well
when there are much fewer number of anomalies than normal data points.
This is called a class imbalance problem.
Since the anomalies rarely occur, it is difficult to collect a sufficient number of them.

% 一方、教師なし手法がこんなとこがいいけど、こんなとこがダメ
On the other hand, an unsupervised approach does not require anomalies for training,
since it tries to model only the normal data points.
Figure \ref{fig:1}(b) shows the image of the unsupervised approach.
Since this approach detects anomalies by using the difference between normal and test data points,
it can detect both known anomalies and unknown anomalies that are away from normal data points.
This difference is frequently called the anomaly score.
To detect anomalies in high-dimensional and complex data, various methods based on deep learning techniques have been presented.
One of the major unsupervised approaches based on deep learning is the autoencoder (AE) \cite{lyudchik2016outlier,sakurada2014anomaly,ma2013parallel}.
The AE is composed of two neural networks: the encoder and the decoder.
The encoder compresses data points into low-dimensional latent representations.
The decoder reconstructs data points from latent representations.
The parameters of the encoder and decoder are optimized by minimizing the error between data points and reconstructed data points,
which is called the reconstruction error.
Since the AE is trained with normal data points,
it becomes able to reconstruct normal data points and fails to reconstruct anomalies.
Therefore, the reconstruction error can be used as the anomaly score.
However, an unsupervised approach performs inferiorly to a supervised approach at detecting known anomalies since it does not utilize the label information.
Furthermore, even if anomalies are obtained in advance, an unsupervised approach cannot utilize these anomalies for improving detection performance.
To handle this problem, the limiting reconstruction capability (LRC), which is the simple extension of the AE, has been presented \cite{munawar2017limiting}.
The LRC maximizes the reconstruction error for the anomaly data points, in addition to minimizing the reconstruction error for normal data points.
As a result, the LRC can detect the known anomalies close to normal data points.
However, the LRC has a serious drawback: the training of AE in the LRC is unstable.
Since the reconstruction error is bounded below but not bounded above,
the LRC tends to maximize the reconstruction error for known anomalies rather than to minimize the reconstruction error for normal data points.
As a result, the LRC cannot reconstruct the normal data points well.

In this paper, we propose the Autoencoding Binary Classifier (ABC), which is a novel supervised approach to exploit the benefits of both supervised and unsupervised approaches.
% In this paper, we propose the Autoencoding Binary Classifiers (ABC), which are a robust supervised anomaly detector based on the AE.
% In the ABC, we assume that label variables follow a Bernoulli distribution.
% The conditional probability of the normal label given a data point is modeled by using the reconstruction error of the AE.
In the ABC, we regard the conditional probability of the label variable given a data point as the Bernoulli distribution.
Its negative log likelihood is modeled by the reconstruction error of the AE.
Thus, by minimizing the negative log likelihood, the AE in the ABC is trained so as to minimize the reconstruction error for normal data points and to maximize the reconstruction error for known anomalies.
Although the behavior of the AE in the ABC is similar to the LRC, the training of the ABC is stable since its objective function is bounded below and above with respect to reconstruction error.
After the training, we can obtain the conditional probability of the label variable given a data point, which is more reasonable anomaly score than the reconstruction error.
As shown in Figure \ref{fig:1}(c), the ABC can detect the anomalies located away from normal data points and detect the anomalies that are close to the known anomalies.
% Since the ABC is based on the Bernoulli distribution, the objective function is bound below and above with respect to reconstruction error, which make the training of the ABC stable.
% Thus, the ABC detect both known and unknown anomalies more accurately than LRC.
% In addition, since the ABC corresponds to the unsupervised approach when there are no known anomalies,
% it can outperform other supervised approaches when a sufficient number of known anomalies are not given in advance.
In addition, even if training dataset does not contain the known anomalies enough, the ABC can outperform the supervised approaches since the ABC behaves as the unsupervised approach.

\section{Preliminaries} \label{sec:ae}  %%%%%%%%%%%%%%%%%%%%%%%%%%%%%%%%%%%%%%%%%%%%%%%%%%%%%%%
First, we review the autoencoder (AE) \cite{hinton2006reducing} and limiting reconstruction capability method (LRC) \cite{munawar2017limiting}.

The AE is a dimensionality reduction algorithm using neural networks.
The AE consists of two parts of neural networks: the encoder and the decoder.
The encoder $E(x)$ compresses data point $x$ into low-dimensional latent representation $z$.
The decoder $D(z)$ reconstructs data points from latent representation $z$.
In the AE, the objective function for each data point is given by
\begin{equation} \label{eq:ae}
  \mathcal{L}_{AE}(x) =  || x - D(E(x)) || ,
\end{equation}
where $||\cdot||$ denotes an arbitrary distance function.
The $\ell_2$-norm is usually used.
This objective function is called the reconstruction error.
The parameters of the encoder and decoder are optimized by minimizing the sum of the reconstruction error for all data points.

One of the most widely used variants of the AE is the Denoising autoencoder (DAE) \cite{vincent2008extracting},
which tries to reconstruct original data points from the noisy input data.
The DAE is estimated by minimizing the following objective function
\begin{equation} \label{eq:dae}
  \mathcal{L}_{DAE}(x) = || x - D(E(x+\epsilon)) || ,
\end{equation}
where $\epsilon$ is a noise from an isotropic Gaussian distribution.
Since the DAE is robust against the noise, it is useful in noisy environments.
The AE and DAE can be optimized with stochastic gradient descent (SGD) \cite{kingma2014adam}.

There are various anomaly detectors based on the AEs \cite{zhou2017anomaly,xie2016unsupervised,zhai2016deep,zong2018deep}.
As a simple way, the reconstruction error can be used for anomaly detection \cite{sakurada2014anomaly}.
If we trained the AE on normal data points, the reconstruction error for anomalies can be larger than that for normal data points. Therefore, we can detect the anomalies by using the reconstruction error as anomaly score.
% , where the AE is trained with normal data points.
% Then, the reconstruction error becomes small for normal data points and large for anomalies.
However, this approach is likely to fail to detect the anomalies near the normal data points.
Furthermore, even if anomalies are obtained in advance, the AE cannot utilize the label information.

% To avoid this, the LRC has been presented.
% The LRC is supervised anomaly detector based on the AE.
The LRC is the simple extension of the AE.
The LRC tries to minimize the reconstruction error of the AE for normal data points and to maximize the reconstruction error for known anomalies.
Suppose that we are given a training dataset $\{ (x_1, y_1), (x_2, y_2), \cdots, (x_N, y_N) \}$, where $x_i$ represents the $i$-th data point, $y_i \in \{ 0, 1 \}$ denotes its label, $y_i = 1$ represents a normal data point and $y_i = 0$ represents an anomaly.
The objective function of the LRC to be minimized is
\begin{align} \label{eq:lrc}
  y_i \mathcal{L}_{AE}(x_i) - (1-y_i)\mathcal{L}_{AE}(x_i),
\end{align}
where $\mathcal{L}(x_i)$ represents the reconstruction error of the AE for data point $x_i$.
By using reconstruction error as anomaly score, the LRC can detect both known and unknown anomalies.
However, the training of the LRC is unstable.
% ここで書かなくてもいいかも。可能なら実験のところに入れたい。
% when a lot of known anomalies included in the training dataset.
% As shown in Eq. (\ref{eq:lrc}), the objective function is not bounded below with respect to reconstruction error.
% Since the reconstruction error bounded below but does not bounded above,
% it is difficult to make the reconstruction error small for even normal data points.
% This divergence may cause poor detection performance.
As shown in Eq. (\ref{eq:lrc}), the LRC tries to maximize the reconstruction error for known anomalies and minimize the reconstruction error for normal data points.
Since the reconstruction error is bounded below but not bounded above, the LRC tends to maximize the reconstruction error for known anomalies rather than to minimize the reconstruction error for normal data points.
As the result, the LRC cannot reconstruct the normal data points well.

% it is assumed that normal data is distributed around the manifold where low dimensional representation is possible.
% Since the normal data occupies most of the training dataset, in order to minimize the overall reconstruction error, the AE will obtain the low dimensional representation of only normal data in hidden layer.
% Thus, if appropriate dimensions reduction is done, the test data around the manifold of normal data should be able to be reconstructed successfully.
% On the other hand, abnormal data points away from the manifold of normal data should fail to reconstruct.
% In other words, this method is based on the empirical observation: the reconstruction error of normal data should be smaller that that of abnormal data not learned.
% Therefore, the reconstruction error itself can be regarded as an anomaly score.

\section{Proposed methods} \label{sec:abc} %%%%%%%%%%%%%%%%%%%%%%%%%%%%%%%%%%%%%%%%%%%%%%%%%%%%%%%
We propose a new supervised anomaly detector based on the AE, the Autoencoding Binary Classifiers (ABC), which can accurately detect known and unknown anomalies.

The ABC uses a probabilistic binary classifier for anomaly detection.
% Suppose that we are given a training dataset $\{ (x_1, y_1), (x_2, y_2), \cdots, (x_N, y_N) \}$ , where $x_i$ represents the $n$-th data point, $y_i \in \{ 0, 1 \}$ denotes its label, $y_i = 1$ represents a normal data point and $y_i = 0$ represents an anomaly.
The probabilistic binary classifier predicts the label $y_i$ from the data point $x_i$.
We assume that the conditional probability of $y_i$ given $x_i$ follows the Bernoulli distribution,
\begin{equation} \label{eq:ber}
  p(y_i|x_i) = \left[ \eta(x_i) \right]^{y_i} \left[ 1 - \eta(x_i) \right]^{1-y_i} ,
\end{equation}
where $\eta(x_i) = p(y_i=1|x_i) = E[y_i|x_i]$ is called the regression function, and its output range is $[0, 1]$.
We can regard $p(y_i=0|x_i)=1-\eta(x_i)$ as the probability that $x_i$ is anomaly.
% Therefore, the regression function $\eta(x_i)$ can be regarded as the anomaly score of input data point $x_i$.
Since the regression function gives low values for known anomalies and high values for normal data
points by maximizing the likelihood of Eq. (\ref{eq:ber}), the ABC can detect known anomalies.
To detect unknown anomalies, we want to make the regression function give low values for unknown anomalies.
The ABC uses the reconstruction error of the AE for the regression function as follows:
\begin{equation}
   \eta_{\theta}(x_i) \equiv e^{-\mathcal{L}_{\theta}(x_i)},
\end{equation}
where $\theta$ is the parameter of the AEs.
The ABC can use the reconstruction error of the AE (Eq. (1)) or the DAE (Eq. (2)) for $\mathcal{L}_{\theta}(x_i)$.
This function takes one when the reconstruction error is zero, and becomes close to zero asymptotically when the reconstruction error goes to infinity.
The range of this regression function is $[0, 1]$.
Since this regression function is based on the AE, it gives low values for unknown anomalies.

\begin{figure}[t]
  \begin{center}
    \includegraphics[width=45truemm]{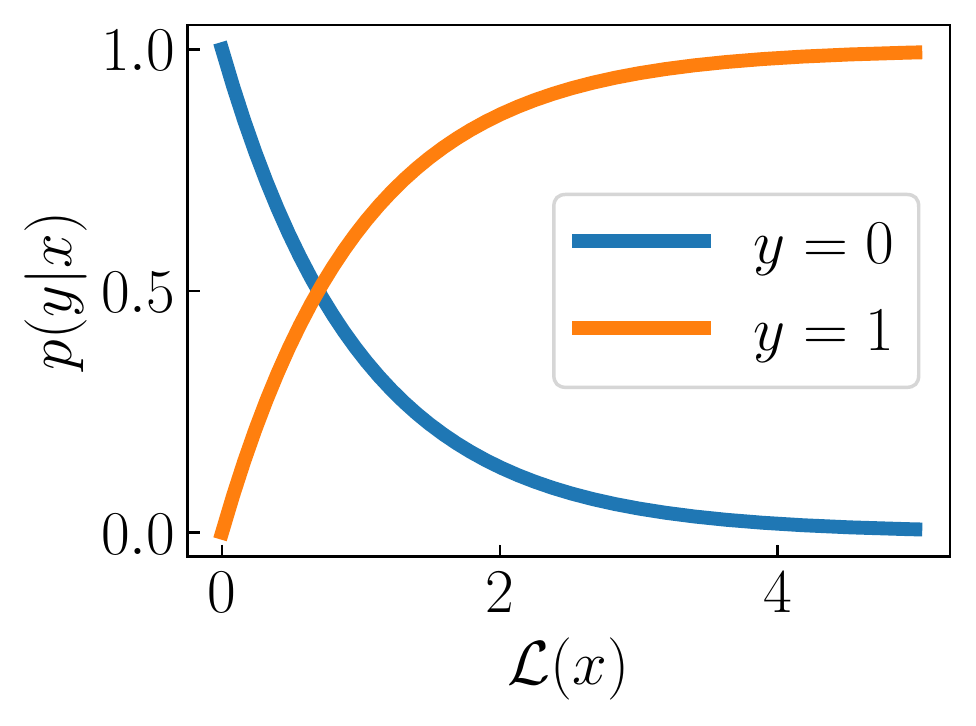}
  \end{center}
  \caption{Shape of objective function of the ABC with respect to reconstruction error $\mathcal{L}_{\theta}(x)$.}
  \label{fig:2}
\end{figure}

Under this definition, we minimize the sum of the following negative log-likelihood of the conditional probability
\begin{align}
  &-\log p_{\theta}(y_i|x_i) \\
  &\quad \!=\! - y_i \log e^{-\mathcal{L}_{\theta}(x_i)} \!-\! (1-y_i) \log (1-e^{-\mathcal{L}_{\theta}(x_i)}) \\
  &\quad \!=\! y_i \mathcal{L}_{\theta}(x_i) \!-\! (1-y_i) \log (1-e^{-\mathcal{L}_{\theta}(x_i)}). \label{eq:obj}
\end{align}
This objective function is equal to that of the AE (Eq. (\ref{eq:ae})) when the input data is normal ($y_i = 1$).
Therefore, if the training dataset consists of only normal data points, this model is identical to the AE.
On the other hand, when the input data is anomaly ($y_i = 0$), this model tries to maximize the reconstruction error $\mathcal{L}_{\theta}(\cdot)$.
The reason is that $ - \log (1-e^{-\mathcal{L}_{\theta}})$ is monotonically decreasing for the reconstruction error $\mathcal{L}_{\theta} \geq 0$ as shown in Figure \ref{fig:2}.
% Note that the reconstruction error will not diverge because the gradient of Eq. (\ref{eq:obj}) decreases as the value of $\mathcal{L}_{\theta}$ increases.
% Therefore, we can stably train this model.
Note that the reconstruction errors for normal and anomaly data points are reasonably balanced by using the likelihood (Eq. (\ref{eq:obj})), and thus, we can stably train the ABC, unlike the LRC.
Furthermore, the ABC also works well for imbalanced data since it exploits the reconstruction error of the unsupervised AE.

% Our ABC regards both the data points away from normal data points and the data points close to known anomalies as anomalies.
% Therefore, it can detect both the known and unknown anomalies accurately in a principled framework that unifies supervised and unsupervised anomaly detection approaches.

Our model can be optimized with SGD by minimizing sum of Eq. (\ref{eq:obj}) for all data points.
Algorithm \ref{alg:proposed} shows pseudo code of the proposed ABC,
where $K$ is the minibatch size and $\theta$ is the parameters of the AE in our model.
After the training, this model can detect anomalies by using the conditional probability $p(y_i|x_i)$ for each data point.
$p(y_i=0|x_i)$ is more reasonable anomaly score than the reconstruction error because we can naturally regard the data point $x_i$ as anomaly when $p(y_i=0|x_i) > 0.5$.
% TODO: できれば、実験結果ものせるとよい。

\begin{algorithm}[t]
  \caption{Autoencoding Binary Classifiers}
  \begin{algorithmic}
    \While{not converged}
      \State{Sample minibatch $\left\{(x_1, y_1), \cdots, (x_K, y_K)\right\}$ from dataset}
      \State{Compute the gradient of $\theta$ with respect to (\ref{eq:obj}):}
      \Statex{$\qquad\qquad g_{\theta} \gets
      \frac{1}{K}\sum_{i=1}^{K} \nabla_{\theta} \left[
      - \log p_{\theta}(y_i|x_i)
      \right]$
      }
      \State{Perform SGD-update for $\theta$ with $g_{\theta}$}
      % \State{Perform SGD-update for $\theta$:}
      % \Statex{$\qquad\qquad \theta \gets \theta + h_i g_{\theta}$}
    \EndWhile
  \end{algorithmic}
  \label{alg:proposed}
\end{algorithm}

\section{Related Work} %%%%%%%%%%%%%%%%%%%%%%%%%%%%%%%%%%%%%%%%%%%%%%%%%%%%%
There is a large literature on anomaly detection.
Here, we briefly review anomaly detection methods by categorizing them into supervised and unsupervised.
In this paper, we define a supervised approach as a method that requires anomalies for training,
an unsupervised approach as a method that can be learned without labeled anomalies.
% , and a semi-supervised approach as a method that uses unlabeled data points to improve the performance of the supervised approach.

If the label information is given perfectly, anomaly detection can be regarded as a binary classification problem.
In this situation, supervised classifiers such as support vector machines \cite{hearst1998support}, gradient tree boosting \cite{friedman2002stochastic} and feed-forward neural networks \cite{dreiseitl2002logistic} are usually used.
However, these standard supervised classifiers cannot detect unknown anomalies accurately and do not work well in the class imbalance situations.
There are several approaches for imbalanced data such as cost-sensitive and ensemble approaches such as random undersampling boost \cite{seiffert2010rusboost} although these approaches do not aim to detect unknown anomalies.
% cost-sensitive approach はcostを手で設定する必要があるってのをコメント
Our ABC also works well for imbalanced data and can detect unknown anomalies since it exploits the reconstruction error of the unsupervised AE.
To achieve high detection performance when label information is available for part of the dataset,
semi-supervised approaches \cite{kiryo2017positive} that utilize both labeled and unlabeled data have been presented.
% In semi-supervised approaches, positive-unlabeled (PU) learning and positive-negative-unlabeled (PNU) learning are the popular methods.
% The PU learning uses normal and unlabeled data \cite{kiryo2017positive},
% and the PNU learning uses the anomalies in addition to them \cite{sakai2016semi}.
The semi-supervised situation that can use the unlabeled data is out of scope of this paper.

Since the label information is rarely given perfectly in real situation, unsupervised approaches such as the local outlier factor \cite{breunig2000lof}, one-class support vector machines \cite{tax2004support}, and the isolation forest method \cite{liu2008isolation} are usually used.
Especially, neural network based methods, typically the AE and
the variational autoencoder (VAE) \cite{kingma2013auto}, have succeeded in detecting anomalies for high-dimensional datasets.
However, the unsupervised approaches do not detect known anomalies as accurately as the supervised approach.
To solve this problems, the LRC has been presented as shown in Section \ref{sec:ae}.
This is similar approach to ours.
However, the training of the LRC becomes unstable since the reconstruction error of the AE in the LRC is bounded below but not bounded above.
Opposite to this, our ABC can be trained stably since the objective function of the ABC is bound above with respect to reconstruction error as described in Section \ref{sec:abc}.

% The PNU learning assume that the normal, anomaly, and unlabeled data can be used for training \cite{sakai2016semi}.
% We exprimentaly found that the ABC can keep the performance in the case of the training data set containing a small amount of unlabeled anomalies.
% We can train our ABC in the case of normal data points containing a small amount of unlabeled anomalies.
% Thus, our proposed ABC can be used in such cases of both PU and PNU anomaly detection settings.

% Supervised versions of the unsupervised AE based methods have been
% presented \cite{munawar2017limiting} that, similarly to our ABC, exploit benefits of both supervised and unsupervised approach.
% We call this method as limiting reconstruction capability (LRC).
% % In \cite{kawachi2018complementary}, a supervised extension of a VAE-based anomaly detector has been presented,
% % which splits the latent space into two classes to detect normal and anomaly data points.
% % As a method similar to the proposed method, the technique to limit the reconstruction capability of the autoencoder using negative learning was presented in \cite{munawar2017limiting}.
% % This method optimizes following objective function instead of our objective Eq. (\ref{eq:obj}).
% % $y_i \mathcal{L}(x_i) - (1-y_i) \mathcal{L}(x_i)$.
% As shown in Sec. \ref{sec:ae}, the training of the LRC tends to become unstable since the reconstruction error in objective function of the LRC may diverge infinitely.
% On the other hand, our ABC is stable since the reconstruction error will not diverge as shown in Figure \ref{fig:2}.

\section{Experiments} %%%%%%%%%%%%%%%%%%%%%%%%%%%%%%%%%%%%%%%%%%%%%%%%%%%%%%%.

\begin{table}[t]
  \begin{center}
    \caption{
      Details of the datasets used in experiments.
      $\mathcal{D}$, $\mathcal{N}$, $\mathcal{A}$, and $\mathcal{U}$ are the numbers of attributes, normal data points, known anomalies, and unknown anomalies, respectively.
    }
    \label{tab:dataset}
    \scalebox{0.9}{
      \begin{tabular}{lrrrr}
      \toprule
                                    & 2D-Toy & MNIST  & KDD`99 & CIFAR10 \\
      \midrule
      $\mathcal{D}$                 & 2       & 784    & 38     & 3,072   \\
      $\mathcal{N}$ (train)         & 10,000  & 17,791 & 87,721 & 5,000   \\
      $\mathcal{A}$ (train)         & 10,000  & 13,797 & 47,869 & 5,000   \\
      $\mathcal{N}$ (test)          & 10,000  & 17,791 & 47,905 & 1,000   \\
      $\mathcal{A}$ (test)          & 10,000  & 13,797 & 17,797 & 1,000   \\
      $\mathcal{U}$ (test)          & 10,000  & 6,824  & 2,926  & 1,000   \\
      \bottomrule
      \end{tabular}
    }
  \end{center}
\end{table}

% In this section, we experimentally show that our approach is effective for detecting both known and unknown anomalies.

\subsection{Datasets}
% イカれたdatasetたちを紹介するぜ
We used the following four datasets.
% We used four datasets: 2D-Toy, MNIST, KDD`99 and CIFAR-10.

\begin{figure*}[!t]
  \begin{center}
    \scalebox{1.0}[1.0]{
      \subfloat[2D-Toy]{
        \includegraphics[width=35truemm]{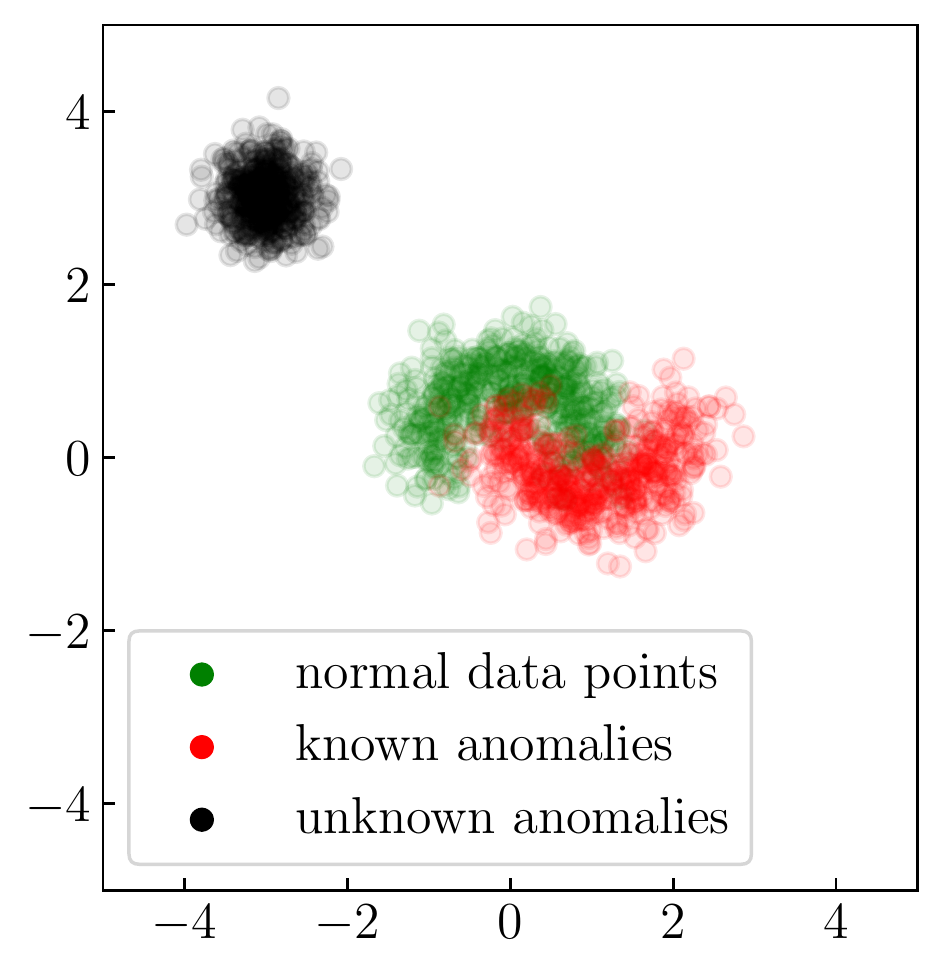}
      }
      \subfloat[DNN]{
        \includegraphics[width=35truemm]{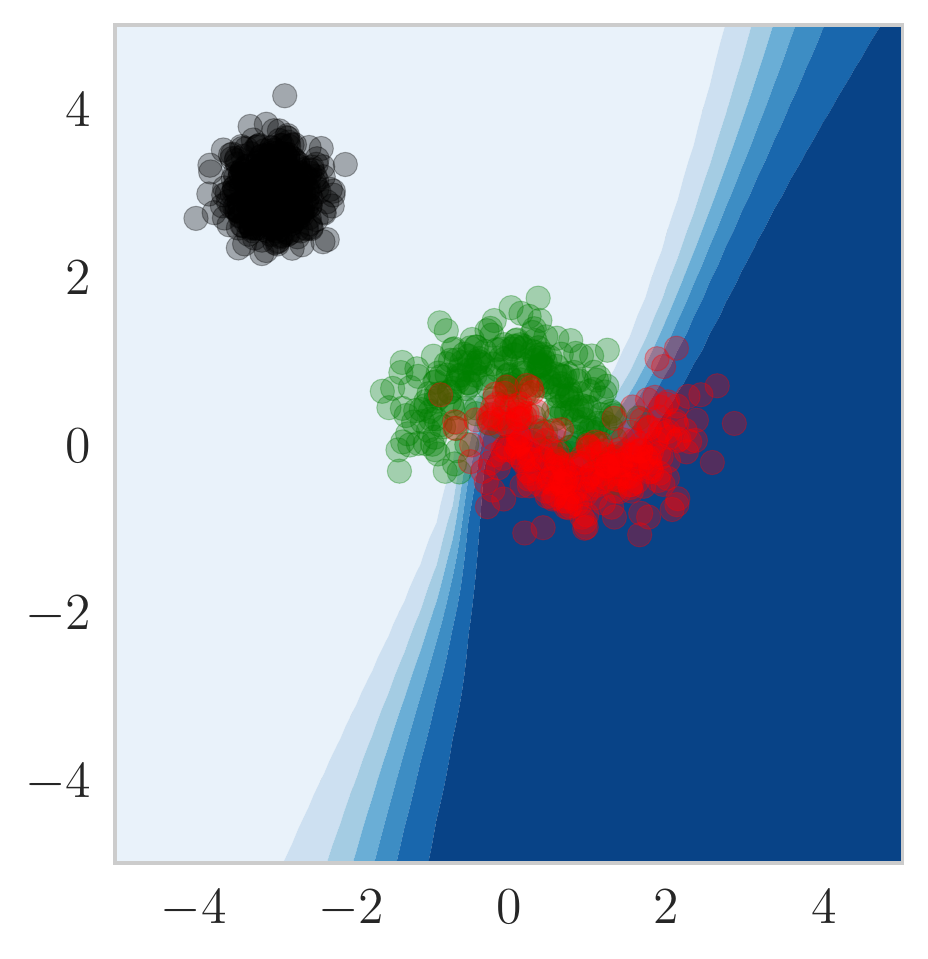}
      }
      \subfloat[Autoencoder]{
        \includegraphics[width=35truemm]{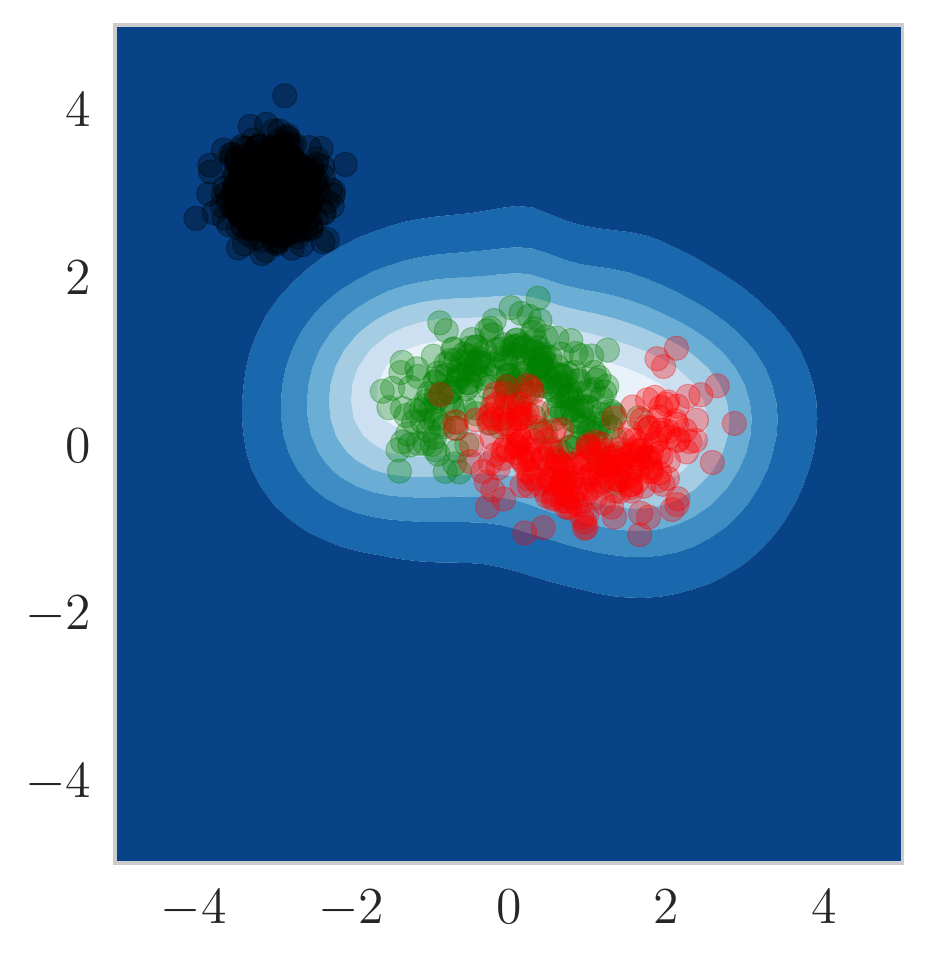}
      }
      \subfloat[ABC (proposed)]{
        \includegraphics[width=35truemm]{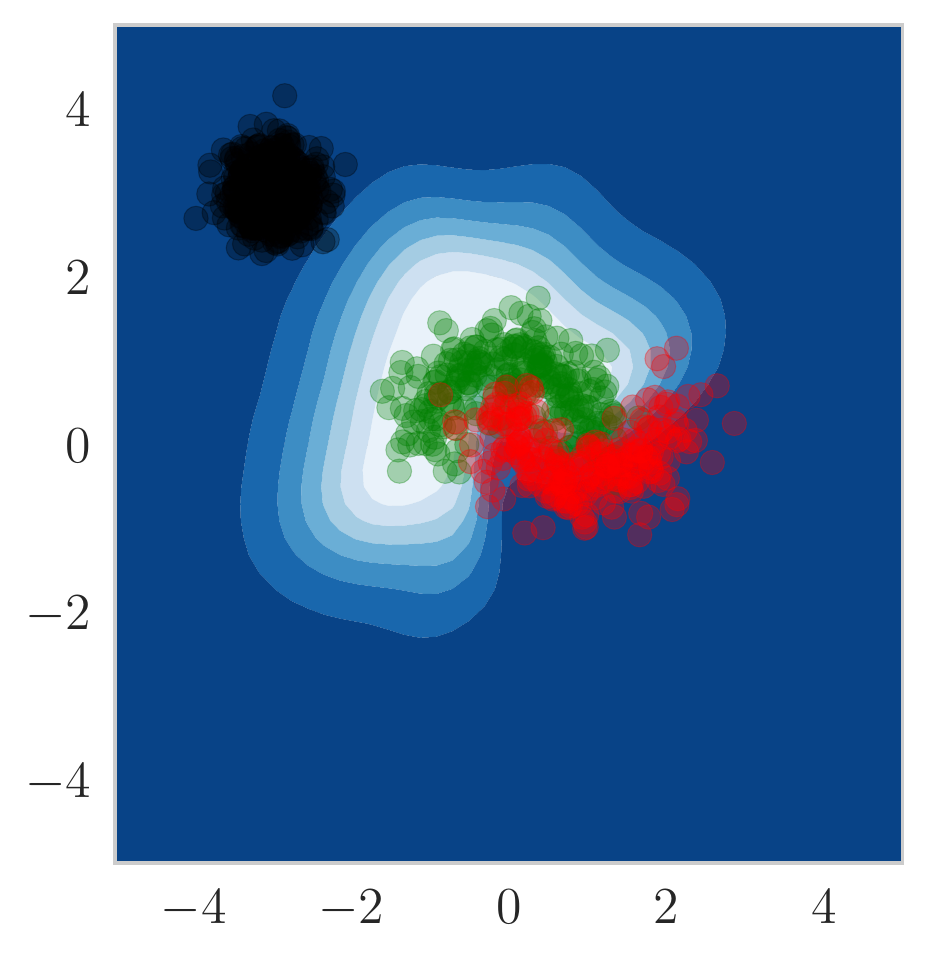}
      }
    }
    \caption{
    (a) Visualization of 2D toy data. Green, red, and black data points represent normal data points, known anomalies, and unknown anomalies, respectively.
    (b)-(d) Heatmap of anomaly scores on 2D-toy data in setting 1.
    }
    \label{fig:heatmap}
  \end{center}
\end{figure*}

% \begin{itemize}
  {\bf 2D-Toy.}
  In order to explain the strengths of our approach, we use the simple two-dimensional dataset consisting of   normal data points, known anomalies that are close to normal data points, and unknown anomalies that are away from normal data points.
  Fig. \ref{fig:heatmap}(a) shows this dataset.
  We generated two interleaving half-circle distribution near the $(0, 0)$.
  We regarded samples from the upper and lower distributions as normal data points and known anomalies, respectively.
  For unknown anomalies, we generated samples from a Gaussian distribution with a $(-3, 3)$ mean and a standard derivation of $0.3$.

  {\bf MNIST.}
  The MNIST consists of hand-written images of digits from $0-9$ \cite{lecun1998gradient}.
  In this data, we used the digits $1, 3, 5, 7, 9$ for normal data points, $0, 2, 6, 8$ for known anomalies, and $4$ for unknown anomalies.
  We used $50\%$ of this dataset for training and the remaining $50\%$ for testing.

  {\bf KDD`99.}
  The KDD‘99 data was generated using a closed network and hand-injected attacks for evaluating the performance of supervised network intrusion detection.
  The dimensionality of the data point was $41$.
  We used `10 percent dataset' for training and `corrected dataset' for test.
  \footnote{This dataset is available at \url{http://kdd.ics.uci.edu/databases/kddcup99/kddcup99.html}}.
  For simplicity, we removed three categorical features (`protocol type', `service' and `flag') and duplicated data points.
  We used `neptune', which is the largest anomaly class, for known anomalies,
  and used attacks included in `R2L' for unknown anomalies.

  {\bf CIFAR10.}
  The CIFAR10 is a collection of images that contains 60,000 32$\times$32 color images in 10 different classes.
  We used `data\_batch\_$1-5$' for training, and `test\_batch' for testing.
  \footnote{This dataset is available at \url{https://www.cs.toronto.edu/~kriz/CIFAR.html}}.
  We used a set of `automobile' images for normal data points, `truck' for known anomalies, and `dog' for unknown anomalies.

% \end{itemize}
The details of datasets are listed in Table \ref{tab:dataset}.
We designed the experiments so that detecting unknown anomalies is more difficult than
detecting known anomalies in unsupervised fashion except for 2D-Toy dataset.
% TODO: なぜ？そしてどうやって？

\subsection{Comparison methods}
% 比較したメソッド一覧はこいつらだ
We compared our approach with the following supervised and unsupervised methods:
% \begin{itemize}

  {\bf Supervised methods.}
  As the supervised approaches, we used the LRC, Support Vector Machine (SVM) classifier \cite{hearst1998support}, Gradient Tree Boosting (GTB) \cite{friedman2002stochastic,chen2016xgboost}, Deep Neural Network (DNN) \cite{dreiseitl2002logistic}, and Random undersampling boost (RUSBoost) \cite{seiffert2010rusboost}.
  For the SVM, we used the radial basis function (rbf) kernel.
  For the DNN, we used a network that had three fully-connected layers and sigmoid cross entropy as a loss function.

  % {\bf Limiting reconstruction capability (LRC).}
  % The LRC is our baseline supervised method as introduced in Section \ref{sec:ae}.
  % % The LRC is a supervised extension of AE-based anomaly detector that limit the reconstruction capability of the AE by using negative learning \cite{munawar2017limiting}.
  %
  % {\bf Support Vector Machine (SVM) classifier.}
  % The SVM classifier is a kernel-based method used for supervised anomaly detection \cite{hearst1998support}.
  % % This method performs the classification by constructing hyperplanes in a multidimensional space that separates cases of different class labels.
  % In these experiments, we used the radial basis function (rbf) kernel for all tasks.
  %
  % {\bf Gradient Tree Boosting (GTB).}
  % The GTB is a tree-based supervised classifier method, that is an ensemble of weak prediction models \cite{friedman2002stochastic,chen2016xgboost}.
  %
  % {\bf Deep Neural Network (DNN).}
  % The DNN is a feedforward neural network for a supervised binary classifier.
  % In these experiments, we used a network that had three fully-connected layers and sigmoid cross entropy \cite{dreiseitl2002logistic} as a loss function.
  %
  % {\bf Random undersampling boost (RUSBoost).}
  % RUSBoost is the supervised method that combine the random undersampling and AdaBoost to achieve high classification performance in class imbalance situations \cite{seiffert2010rusboost}.

  {\bf Unsupervised methods.}
  As the unsupervised approaches, we used the AE, DAE, One-class SVM (OCSVM) \cite{tax2004support}, and Isolation Forest (IF) \cite{liu2008isolation}.
  For the OCSVM, we used the rbf kernel.

  % {\bf Autoencoder (AE).}
  % The AE is our baseline unsupervised method as introduced in Section \ref{sec:ae}.
  % % We regarded the reconstruction error as anomaly scores for each data point.
  % We also tested the Denoising AE (DAE).
  %
  % {\bf One-class SVM (OCSVM).}
  % The OCSVM is the SVM variant for unsupervised anomaly detection \cite{tax2004support}.
  % % The OCSVM is a popular kernel-based unsupervised algorithm that learns a decision function for anomaly detection.
  % In these experiments, we used the rbf kernel for all tasks.
  %
  % {\bf Isolation Forest (IF).}
  % The IF is a tree-based unsupervised anomaly detection method \cite{liu2008isolation}.
  % This method isolates anomalies by randomly selecting features and randomly choosing split values between the maximum and minimum of the selected features.

  % {\bf Semi-supervised methods.}
  % As the semi-supervised approach, we used the Non-Negative Positive-Unlabeled learning (nnPU) \cite{kiryo2017positive}.
  % We used a three-layer perceptron as the classifier and sigmoid as the loss function.

% \end{itemize}
% 実装にはあれを用いたぜって書けたら書く

\subsection{Setup}
% データを3つにわけてうんぬん
We did experiments in following three settings:

  {\bf Setting 1.}
  To evaluate detection performance for both known and unknown anomalies,
  we divided datasets into three subsets: normal data points, known anomalies and unknown anomalies.
  The normal data points and known anomalies were used for training supervised methods,
  and only normal data points were used for training unsupervised methods.
  The unknown anomalies were only used for testing.
  In this setting, we regard the unknown anomalies as novel anomalies.

  {\bf Setting 2.}
  This is the same as setting 1 except that $100$ unknown anomalies were included in the normal data points.
  This is the more realistic setting than setting 1.
  The remaining unknown anomalies were used for test.
  % In this setting, we regarded the unknown anomalies as contaminated normal class examples.

  {\bf Setting 3.}
  To measure the detection performances of supervised approaches in class imbalance situations,
  we evaluated their performances when the number of known anomalies is changed on MNIST dataset in setting 1.
  % Note that we tested the nnPU in positive-unlabeled situation; that is, known anomalies that were removed in this setting were used as unlabeled data for training.
  % We also test the nnPU, LRC and our ABC in Positive-Unlabeled situation; that is, reduced known anomalies were included in training dataet as unlabeled data.
We evaluated the detection performance by using the area under the receiver operating characteristic curve (AUROC).
% We evaluated the detection performance by using two metrics:
% the area under the receiver operating characteristic curve (AUROC) and the area under the precision-recall curve (AUPRC). Both metrics are for ranking performance.
% The AUROC is suitable for evaluating the detection performance in general.
% The AUPRC is suitable for class imbalance situations \cite{davis2006relationship}.

% 使用したネットワーク構造はうんぬん
For our approach, we used neural networks with two hidden layer ($10-10$ hidden units for 2D-Toy, and $300-100$ hidden
units for other datasets) for the encoder and the decoder.
We used $\ell_2$-norm as the distance function of Eq. (\ref{eq:ae}) and Eq. (\ref{eq:dae}).
We also evaluated our ABC with the DAE with noise from a Gaussian distribution with a standard deviation of $0.2$.
We used a hyperbolic tangent for the activation function.
As the optimizer, we used Adam with batch size 100 \cite{kingma2014adam}.
The number of latent variables was changed in accordance with the datasets: we set the
dimension of z to one for 2D-Toy and the dimension of $z$ to $20$ for the other datasets.
The maximum number of epochs was set to $300$ for all datasets, and we
used early-stopping on the basis of the validation data \cite{prechelt1998automatic}.
We use same network architecture for the AE, DAE and LRC.
We used $20\%$ of the training dataset as validation data.
Each attribute was linearly normalized into a range $[0, 1]$ by min-max scaling expect for 2D-Toy.
We ran all experiments five times each.

\subsection{Results}
Figures \ref{fig:heatmap}(b)-(d) show the heatmap of anomaly scores of 2D-Toy by the DNN, the AE and our ABC in setting 1.
Table \ref{tab:result} show the AUROC with the proposed method and comparison methods in setting 1 and 2.
We used bold to highlight the best result and the results that are not statistically different from the best result according to a pair-wise $t$-test.
We used $5\%$ as the p-value.

% まずは教師あり学習から
First, we focus on the supervised approaches: DNN, SVM, GTB, and RUSBoost.
Figure \ref{fig:heatmap}(b) shows the heatmap of anomaly scores by the DNN, which achieved the highest performance among supervised approaches for detecting known anomalies.
This result indicates that supervised approaches can discriminate known anomalies from normal data points accurately.
However, these approaches failed to detect unknown anomalies on the opposite side of the known anomalies.
Tables \ref{tab:result} also show that these approaches can detect known anomalies but not unknown anomalies accurately on various datasets\footnote{Note that although the SVM classifier could detect unknown anomalies among supervised approaches, it performed inferiorly to our ABC on average.}.
In contrast to these approaches, our ABC can detect known anomalies as accurately as supervised approaches and unknown anomalies better than these approaches.

% つぎは教師なし学習
Second, we focus on the unsupervised approaches: AE, DAE, OCSVM, and IF.
Figure \ref{fig:heatmap}(c) shows the heatmap of anomaly scores by the AE, which achieved the highest performance among unsupervised approaches.
In contrast to supervised approaches, unsupervised approaches can detect unknown anomalies that are located away from normal data points.
However, these approaches cannot detect known anomalies as accurately as supervised approaches:
they are likely to fail to detect anomalies close to normal data points since these approaches use the difference between normal and observed data points for detection.
Table \ref{tab:result} also show that these approaches can detect unknown anomalies but not known anomalies accurately.
Furthermore, for CIFAR-10 datasets, these approaches perform poorly because this dataset is most complicated and heavily overlapping dataset in our experiments.
% TODO: 理由を言う。CIFAR10はcomplexでheavy overlapedなデータだから、みたいな。mnistとの比較旧でかくこと。
Meanwhile, our ABC achieved high detection performances for both known and unknown anomalies on all datasets.

\begin{figure}[t]
  \begin{center}
    \subfloat[ABC]{
      \includegraphics[width=40truemm]{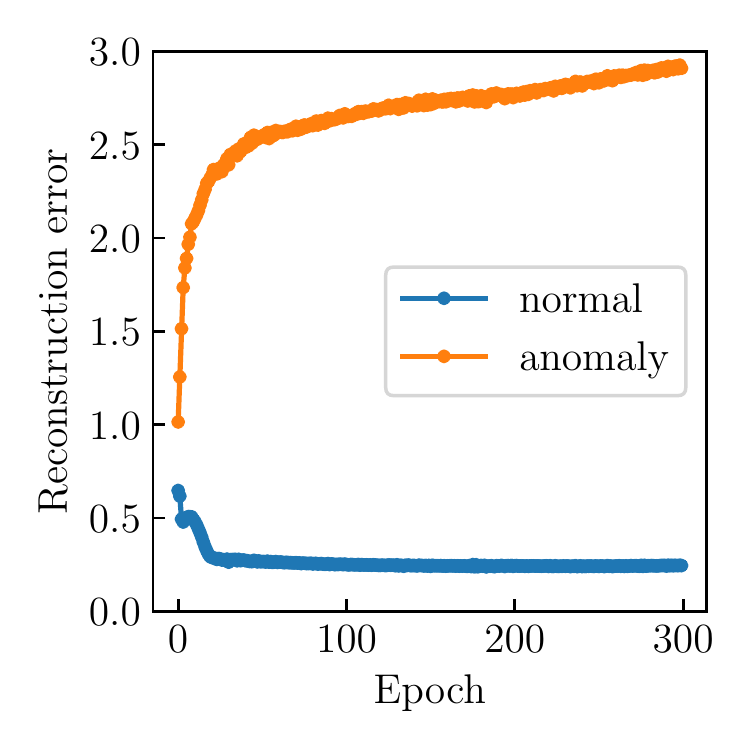}
    }
    \subfloat[LRC]{
      \includegraphics[width=40truemm]{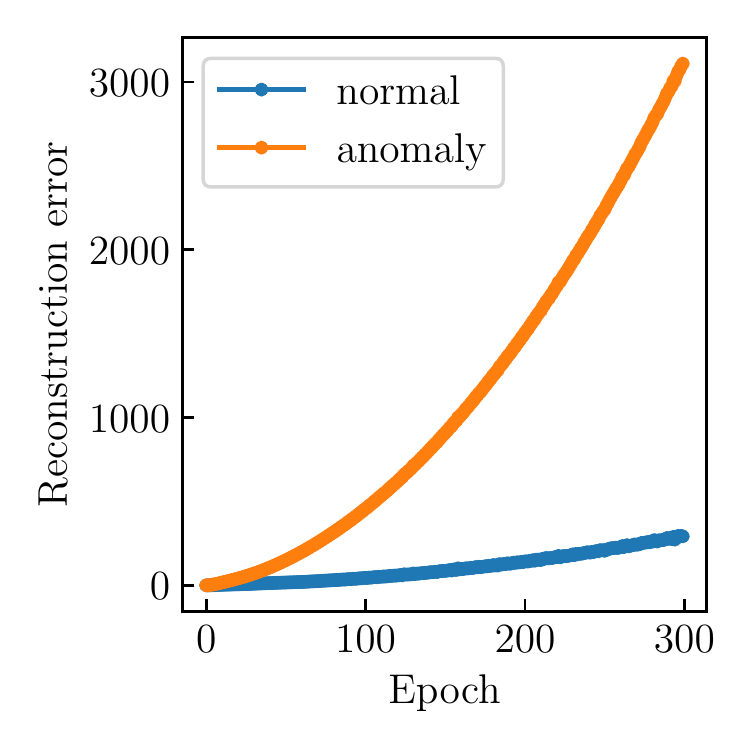}
    }
  \caption{
    Mean training loss for normal data points and anomalies on the 2D-Toy dataset.
  }
  \label{fig:loss}
  \end{center}
\end{figure}

% その他の学習
Third, we focus on the LRC.
% The nnPU learning shows the same performance tendency as the supervised approaches.
% The nnPU learning cannot detect unknown anomalies expect for the KDD`99 dataset.
% This is because nnPU learning aims to provide superior performance on partially labeled datasets and does not directly aims to detect unknown anomalies.
The LRC shows the same performance tendency as our ABC.
The LRC can detect both known and unknown anomalies.
However, the variance of performance is larger than that of the ABC and sometimes it cannot detect both known and unknown anomalies at all.
Furthermore, it performed inferiorly to our ABC in all situations.
To compare the stability of training of the LRC and ABC, we plotted in Figure \ref{fig:loss} the reconstruction error for normal data points and the known anomalies on 2D-Toy dataset.
Figure \ref{fig:loss}(a) shows the mean reconstruction error of the LRC.
As the training proceeds the LRC does not reconstruct normal data points.
The reason is the LRC tends to maximize the reconstruction error for known anomalies rather than to minimize the reconstruction error for normal data points.
This result indicates that the training of the AE in the LRC is unstable.
Opposite to this, the ABC can minimize the reconstruction error for normal data points and maximize the reconstruction error for known anomalies as show in Figure \ref{fig:loss}(b).
This indicates the training of the AE in the ABC is stable.

% 最後にProposed method
Fourth, we focus on our ABC.
Figure \ref{fig:heatmap}(d) shows the heatmap of anomaly scores by our ABC.
In contrast to the AE, the anomaly scores by our ABC are low for normal data points and high for known anomalies.
The reason is that we train the AE in our ABC to minimize the reconstruction error for normal data points and maximize the reconstruction error for known anomalies.
On the other hand, the anomaly scores of our ABC for unknown anomalies are also high since our approach fails to reconstruct data points not included in the training dataset.
% TODO: 何と比べて？
% the reconstruction error becomes large for data points that are not included in training dataset.
Therefore, our ABC can detect both known and unknown anomalies accurately.
Table \ref{tab:result} shows that our ABC achieved the almost equal to or better performance than the other approaches.
These results indicate that our ABC is useful in various tasks.

% TODO: 標準偏差の追加
\begin{table*}[!t]
  \begin{center}
    \caption{Summary of the experiment results AUROC.
    The standard deviation is in parentheses.
    For convenience of space, it is written up to 3 decimal points (e.g., $0.1234$ is expressed as $123$).
    % we reduced the value of the standard deviation less than $10^{-4}$ is expressed as $000$.
    $\mathcal{A}$, and $\mathcal{U}$ represents known anomalies and unknown anomalies, respectively.}
    \label{tab:result}
    \scalebox{0.72}[0.72]{
      \begin{tabular}{llrrrrrrr|rrrr}
        \toprule
        {\bf Setting 1}           &               & ABC(AE)          & ABC(DAE)        & LRC             & DNN             & SVM             & GTB             & RUSBoost        & AE               & DAE             & OCSVM           & IF              \\
        \midrule
        \multirow{2}{*}{2D-Toy}   & $\mathcal{A}$ & 0.965(004)       & 0.966(001)      & 0.914(002)      & {\bf 0.972(000)}& 0.970(000)      & 0.972(000)      & 0.961(002)      & 0.859(005)       & 0.888(005)      & 0.833(000)      & 0.895(006)      \\
                                  & $\mathcal{U}$ & {\bf 1.000(000)} & 0.999(000)      & 0.857(044)      & 0.000(001)      & 0.907(000)      & 0.005(000)      & 0.006(002)      & {\bf 1.000(000)} & {\bf 1.000(000)}& {\bf 1.000(000)}& {\bf 1.000(000)}\\ \cmidrule{2-13}
        \multirow{2}{*}{MNIST}    & $\mathcal{A}$ & {\bf 0.998(000)} & {\bf 0.999(000)}& 0.972(006)      & 0.998(000)      & 0.985(000)      & 0.991(000)      & 0.958(001)      & 0.968(001)       & 0.970(001)      & 0.805(001)      & 0.817(014)      \\
                                  & $\mathcal{U}$ & 0.835(030)       & 0.856(016)      & 0.630(055)      & 0.758(033)      & 0.705(009)      & 0.704(009)      & 0.663(041)      & {\bf 0.892(011)} &{\bf 0.886(010)} & 0.623(002)      & 0.662(012)      \\ \cmidrule{2-13}
        \multirow{2}{*}{KDD`99}   & $\mathcal{A}$ & {\bf 1.000(000)} & {\bf 1.000(000)}& 0.999(001)      & {\bf 1.000(000)}& {\bf 1.000(000)}& 1.000(000)      & {\bf 1.000(000)}& 0.998(002)       & 0.998(001)      & 1.000(000)      & 0.999(001)      \\
                                  & $\mathcal{U}$ & {\bf 0.884(017)} & 0.788(013)      & 0.781(018)      & 0.624(180)      & {\bf 0.893(000)}& 0.621(000)      & 0.735(044)      & 0.841(041)       & 0.793(009)      & 0.762(000)      & 0.842(015)      \\ \cmidrule{2-13}
        \multirow{2}{*}{CIFAR-10} & $\mathcal{A}$ & {\bf 0.847(008)} &{\bf 0.842(010)} & 0.658(123)      & 0.780(006)      & 0.788(000)      & 0.826(000)      & 0.656(027)      & 0.534(002)       & 0.533(001)      & 0.490(000)      & 0.498(010)      \\
                                  & $\mathcal{U}$ & {\bf 0.793(028)} &{\bf 0.785(026)} & 0.624(097)      & 0.613(041)      & 0.550(000)      & 0.628(000)      & 0.557(052)      & 0.465(003)       & 0.465(002)      & 0.495(000)      & 0.531(009)      \\
        \midrule
        {\bf Setting 2}           &               & ABC(AE)         & ABC(DAE)        & LRC             & DNN            & SVM            & GTB            & RUSBoost       & AE              & DAE              & OCSVM          & IF              \\
        \midrule
        \multirow{2}{*}{2D-Toy}   & $\mathcal{A}$ & 0.965(002)      & 0.967(001)      &{\bf 0.836(171)} &{\bf 0.972(000)}& 0.970(000)     & 0.972(000)     & 0.964(001)     & 0.839(026)      & 0.848(040)       & 0.833(000)     & 0.893(006)      \\
                                  & $\mathcal{U}$ &{\bf 0.988(019)} & 0.997(001)      & 0.762(125)      & 0.000(001)     & 0.755(000)     & 0.001(000)     & 0.001(001)     &{\bf 0.767(214)} &{\bf 0.750(228)}  &{\bf 1.000(000)}&{\bf 1.000(000)} \\ \cmidrule{2-13}
        \multirow{2}{*}{MNIST}    & $\mathcal{A}$ & 0.999(000)      &{\bf 0.999(000)} & 0.972(004)      & 0.998(000)     & 0.985(000)     & 0.991(000)     & 0.955(002)     & 0.969(002)      & 0.970(001)       & 0.805(001)     & 0.821(008)      \\
                                  & $\mathcal{U}$ & 0.799(013)      & 0.823(016)      & 0.645(058)      & 0.708(059)     & 0.685(009)     & 0.689(013)     & 0.655(031)     & {\bf 0.862(011)}& {\bf 0.857(010)} & 0.619(002)     & 0.660(013)      \\ \cmidrule{2-13}
        \multirow{2}{*}{KDD`99}   & $\mathcal{A}$ & {\bf 1.000(000)}& {\bf 1.000(000)}& 1.000(000)      &{\bf 1.000(000)}&{\bf 1.000(000)}& 1.000(000)     &{\bf 1.000(000)}& 0.998(000)      & 0.998(001)       & 1.000(000)     & 1.000(000)      \\
                                  & $\mathcal{U}$ & 0.845(019)      & 0.781(016)      & 0.786(040)      & 0.598(185)     &{\bf 0.892(001)}& 0.622(001)     & 0.763(049)     & 0.844(030)      & 0.794(010)       & 0.762(000)     & 0.840(009)      \\ \cmidrule{2-13}
        \multirow{2}{*}{CIFAR-10} & $\mathcal{A}$ & {\bf 0.836(009)}& {\bf 0.844(004)}& 0.602(138)      & 0.780(005)     & 0.787(000)     & 0.823(002)     & 0.660(013)     & 0.532(002)      & 0.531(001)       & 0.492(000)     & 0.503(16)       \\
                                  & $\mathcal{U}$ & {\bf 0.727(029)}&{\bf 0.700(031)} & 0.544(070)      & 0.575(024)     & 0.521(005)     & 0.600(007)     & 0.514(015)     & 0.458(002)      & 0.458(002)       & 0.494(003)     & 0.533(014)      \\
        \bottomrule
      \end{tabular}
    }
  \end{center}
\end{table*}

%%%%%%%%%%%%%%%%%%%%%%%%%%%%%%%%%%%%%%%%%%%%%%%%%%%%%%%%%%%%%%%%%%%%%%%%%%%%%%%%
\begin{figure}[tb]
  \begin{center}
    \includegraphics[width=70truemm]{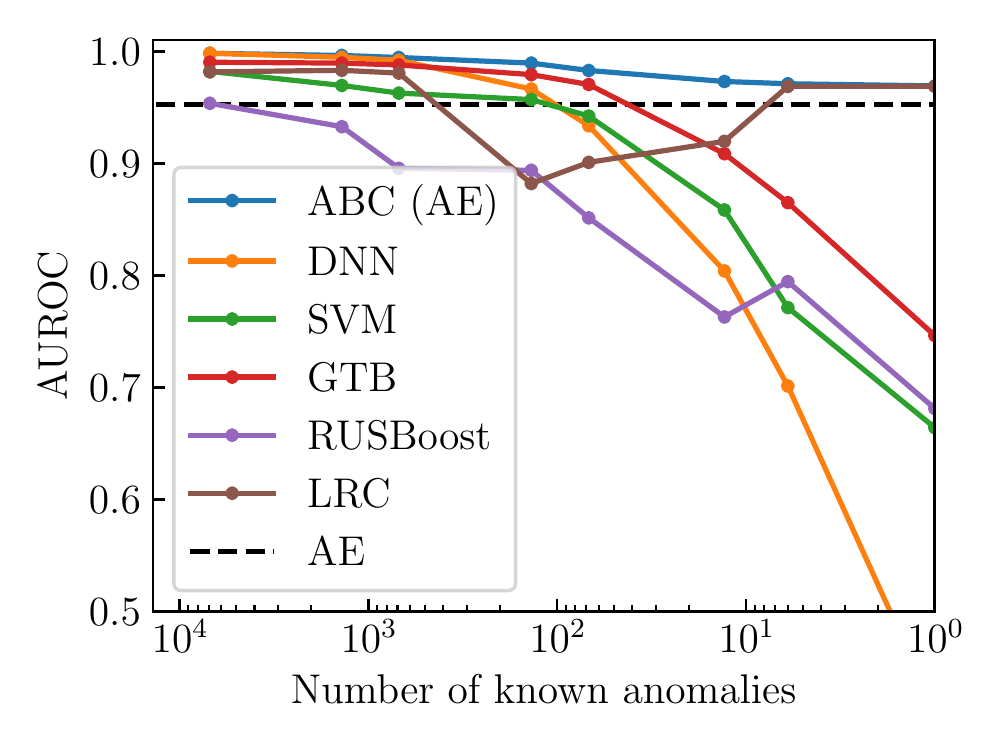}
  \end{center}
  \caption{
    Detection performance comparison when number of known anomalies is reduced.
    For comparison, the dashed line plots the AUROC of the AE that does not use any known anomalies for training.
  }
  \label{fig:imbalance}
\end{figure}

Figures \ref{fig:imbalance} shows the relationship between the AUROC and the number of known anomalies for training in setting 3.
Our ABC maintains a high detection performance, whereas the other supervised approaches perform poorly as the number of known anomalies decreases, even RUSBoost, which is designed to work well in class imbalance situations.
% The semi-supervised learning nnPU shows the same performance tendency as the supervised approaches.
% % The nnPU learning cannot detect unknown anomalies expect for the KDD`99 dataset.
% This is because nnPU learning aims to work well on partially labeled datasets and does not aims to work well in class imbalance situations.
% TODO: 提案手法がclass imbalanceでうまくいく設計になっている理由をsec.3に追加できればする
This result indicates that our ABC is useful when we cannot obtain enough anomalies.

These results indicate that our approach is a good alternative to other approaches: our approach can detect both known and unknown anomalies accurately, and it works well when the number of known anomalies are not enough.

\section{Conclusion}

In this paper, we introduced a new anomaly detector Autoencoding Binary Classifiers (ABC).
We modeled normal data points with the unsupervised approach, and made it fail to model known anomalies by the supervised approach.
We assumed that labels follow a Bernoulli distribution, and modeled its negative log likelihood of the normal label given a data point by using the reconstruction error of the AE.
By maximizing the likelihood of the conditional probability, our AE was trained so as to minimize the reconstruction error for normal data points, and to maximize the reconstruction error for known anomalies.
Since the proposed method becomes able to reconstruct normal data points and does not reconstruct known and unknown anomalies, it can detect both known and unknown anomalies accurately.
The training of the ABC is stable and detection performance is higher than the LRC.
In addition, since our approach corresponds to the AE when there are no known anomalies, our approach work well when we do not obtain the sufficient number of known anomalies in advance.
We experimentally shows that
% the effectiveness of the proposed method.
% Our experimental results indicate that our approach is a good alternative to other approaches:
our approach can detect both known and unknown anomalies accurately and it works well even when the number of known anomalies is not enough.
In the future, we will try to apply our approach to various anomaly detection tasks such as network security and malware detection.

\bibliographystyle{named}
\bibliography{ijcai19}

\end{document}